\documentclass[10pt,twocolumn,letterpaper]{article}

\usepackage[pagenumbers]{cvpr} 

\usepackage{amsmath,amsfonts,bm,mathtools}

\def\eqref#1{equation~\ref{#1}}

\def\1{\bm{1}}

\def\rvepsilon{{\boldsymbol{\epsilon}}}

\def\rvc{{\mathbf{c}}}

\def\rvx{{\mathbf{x}}}

\def\rvz{{\mathbf{z}}}

\def\vc{{\bm{c}}}

\def\mI{{\bm{I}}}

\DeclareMathAlphabet{\mathsfit}{\encodingdefault}{\sfdefault}{m}{sl}
\SetMathAlphabet{\mathsfit}{bold}{\encodingdefault}{\sfdefault}{bx}{n}

\def\gN{{\mathcal{N}}}

\def\@onedot{\ifx\@let@token.\else.\null\fi\xspace}

\newcommand{\E}{\mathbb{E}}

\newcommand{\fullname}{DiffusionHarmonizer\xspace}

\newcommand{\myparagraph}[1]{\vspace{2mm}\noindent\textbf{#1}\,}

\usepackage{microtype}

\newcommand\blfootnote[1]{%
  \begingroup
  \renewcommand\thefootnote{}\footnote{#1}%
  \addtocounter{footnote}{-1}%
  \endgroup
}
\definecolor{cvprblue}{rgb}{0.21,0.49,0.74}
\usepackage[pagebackref,breaklinks,colorlinks,allcolors=cvprblue]{hyperref}

\usepackage{multirow} 

\title{\fullname{}: Bridging Neural Reconstruction and Photorealistic Simulation with Online Diffusion Enhancer}

\author{
Yuxuan Zhang$^1$\footnotemark[1]   \and 
Katar\'ina T\'othov\'a$^1$\footnotemark[1] \and 
Zian Wang$^{1, 2}$ \and 
Kangxue Yin$^1$ \and
Haithem Turki$^1$ \and 
Riccardo de Lutio$^1$ \and 
Yen-Yu Chang$^{1, 3}$ \and 
Or Litany$^{1, 4}$ \and
Sanja Fidler$^{1, 2}$ \and 
Zan Gojcic$^1$ \\ 
\\[-4mm]
\small {$^1$NVIDIA \quad $^2$University of Toronto \quad $^3$Cornell University \quad $^4$Technion} \\
\small\texttt{\{alezhang, ktothova, zgojcic\}@nvidia.com}\\
}

\begin{document}

\twocolumn[{
\maketitle
\renewcommand\twocolumn[1][]{#1}
\begin{center}
    \centering
    \vspace{-8mm}
    \includegraphics[trim={0 10 0 0},clip, width=0.98\linewidth]{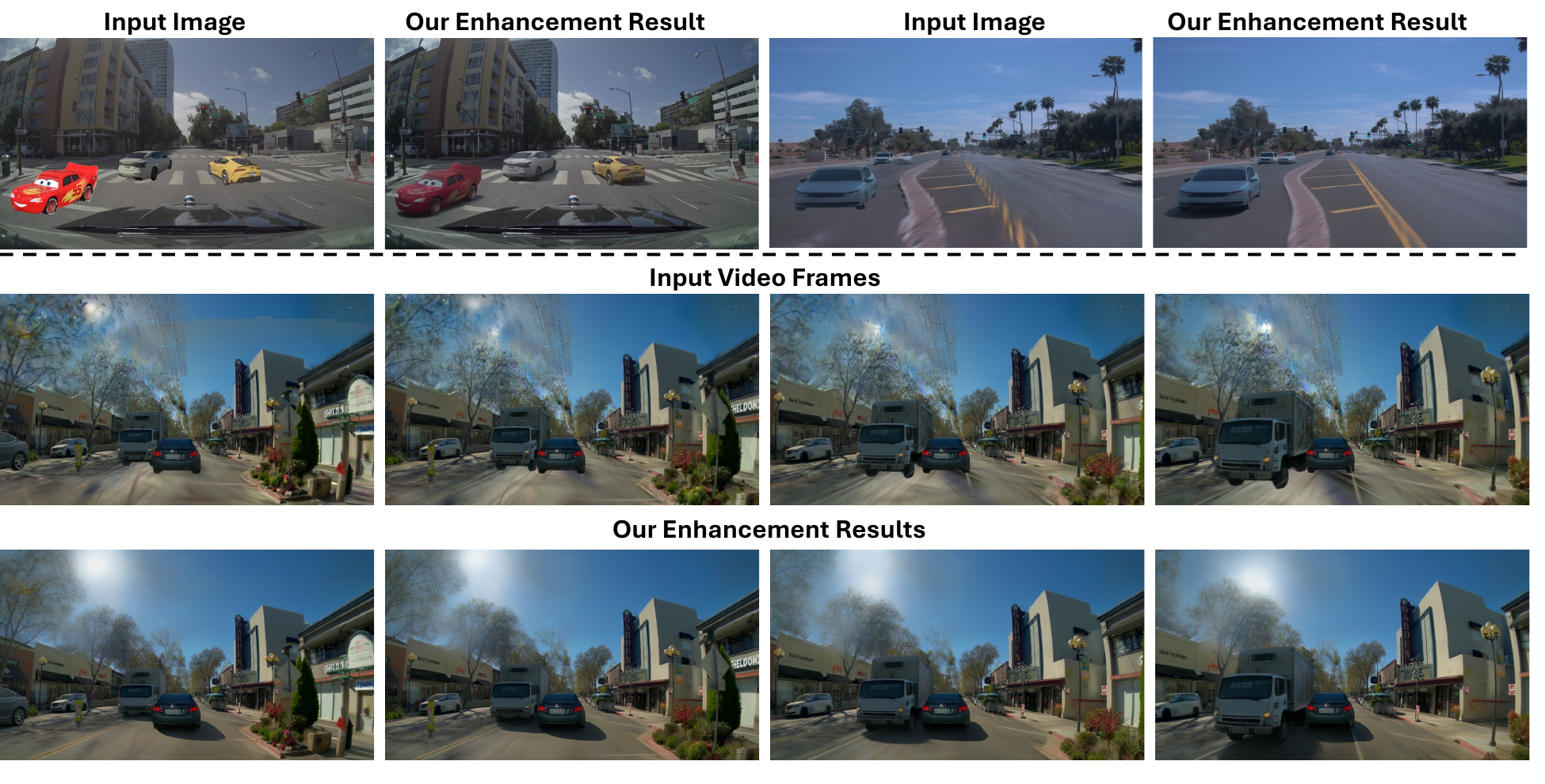}
    \vspace{-3 mm}
    \captionof{figure}{\textbf{\fullname{} on Driving Scenes}. Our method transforms artifact-prone neural-rendered frames into temporally coherent simulations, improving their realism by jointly correcting shadows, lighting, appearance discrepancies and reconstruction artifacts. 
    }
    \label{fig:teaser}
\end{center}
}]

\blfootnote{\hspace{-1em} $^*$ Equal contribution.}

\vspace{-3mm}
\begin{abstract}
Simulation is essential to the development and evaluation of autonomous robots such as self-driving vehicles. 
Neural reconstruction is emerging as a promising solution as it enables simulating a wide variety of scenarios from real-world data alone in an automated and scalable way. 
However, while methods such as NeRF and 3D Gaussian Splatting can produce visually compelling results, they often exhibit artifacts particularly when rendering novel views, and fail to realistically integrate inserted dynamic objects, especially when they were captured from different scenes.
To overcome these limitations, we introduce \fullname{}, an online generative enhancement framework that transforms renderings from such imperfect scenes into temporally consistent outputs while improving their realism.
At its core is a single-step temporally-conditioned enhancer that is converted from a pretrained multi-step image diffusion model, capable of running in online simulators on a single GPU. 
The key to training it effectively is a custom data curation pipeline that constructs synthetic–real pairs emphasizing appearance harmonization, artifact correction, and lighting realism. 
Experiments show that \fullname{} substantially improves perceptual realism, being chosen by $84.28\%$ of users in our comparative study over the second-best method. Furthermore, it matches the temporal coherence of state-of-the-art video models while maintaining the inference efficiency of single-step image models, offering a scalable and practical solution for simulation in both research and production settings.

\end{abstract}    

\vspace{-4mm}
\section{Introduction}
\vspace{-1mm}

\label{sec:intro}

Recent advances in neural reconstruction have made it possible to recover high-fidelity simulation environments directly from real-world sensor data, paving the way for scalable and photorealistic simulation in autonomous driving and robotics~\cite{zhou2024hugsim, chen2025omnire, turki2025simuli, Escontrela25arXiv_GaussGym}. These methods typically decompose a scene into a static background and a set of foreground assets (\eg cars and people) whose positions and trajectories can be manipulated during simulation. The extracted foreground assets can be stored in a databank and flexibly combined with different backgrounds or scenes.

Despite this progress, two fundamental challenges remain: \emph{(1) Novel-view artifacts:} while current methods achieve high visual quality near training viewpoints, they often produce spurious geometry, missing regions, and other artifacts when rendered from sparsely observed or extrapolated viewpoints that significantly deviate from the training trajectory. Similar artifacts arise when the position or trajectory of the foreground assets is manipulated; and \emph{(2) Object insertion artifacts:} when (dynamic) foreground objects, whether synthetic assets or reconstructions from separate captures, are inserted into reconstructed scenes, the resulting composites frequently contain tone discrepancies, missing shadows, or lighting mismatches (see \cref{fig:teaser}).

In this work, we aim to improve the photorealism of neurally reconstructed simulation environments by leveraging generative models as \emph{post-rendering enhancers} to address the above challenges. We formulate this problem as an image-to-image translation task: given artifact-prone or visually inconsistent frames rendered from a reconstructed scene, our goal is to generate temporally consistent and harmonized video frames that preserve the underlying scene structure. 

Although image-to-image translation and video editing have been extensively studied, existing models fail to meet the requirements of online simulation: Video-based generative models are computationally expensive and cannot operate in online simulation under practical resource budgets (\eg, a single H100 GPU), while image-based models lack temporal consistency, leading to flickering and unstable dynamics. 
Both approaches often struggle to reliably model lighting such as cast shadows and may distort existing scene geometry and appearance, which is undesirable for physically grounded simulation.

To address these challenges, we convert a pretrained non-distilled image diffusion model into a single-step, temporally conditioned enhancement model suitable for online simulation. 
This requires a targeted training strategy that preserves scene structure and mitigates artifacts arising from the noise-trajectory mismatch which emerges due to the discrepancy between the multi-step noise schedule used during pretraining and the single-step mode at inference time. Indeed, naively fine-tuning a pretrained multi-step diffusion model in a single denoising step introduces high-frequency checkerboard artifacts. To address this issue, we introduce a multi-scale perceptual loss that stabilizes high-frequency behavior and effectively suppresses artifacts induced by the mismatched denoising trajectories. 

To overcome the lack of high-quality paired supervision, we introduce a scalable data curation pipeline that synthesizes training pairs capturing harmonization, shadow correction, and artifact correction. Specifically, we combine (i) artifact-corrupted renderings generated using the four degradation modes from DIFIX3D+ \cite{difix3d}: sparse reconstruction, cycle reconstruction, cross-referencing, and underfitting; (ii) appearance-varying data created by randomized modifications of ISP parameters such as tone mapping, exposure, and white balance, \etc; (iii) pairs of images with varying illumination synthesized using a generative relighting model~\cite{{DiffusionRenderer}}; (iv) physically based shadow data obtained by rendering synthetic scenes under varying environment-map and light-source configurations; and (v) asset re-insertion composites where reconstructed objects are reinserted without shadows to create realistic, supervision-rich pairs for both harmonization, artifact correction, and shadow synthesis. These components jointly supply the complementary signals required to learn robust appearance harmonization, artifact correction, and physically plausible shadow generation.

By combining this data generation pipeline with our targeted training strategy, we obtain \fullname{}, a unified simulation harmonizer that jointly (i) corrects novel-view reconstruction artifacts, (ii) harmonizes foreground and background appearance, and (iii) synthesizes realistic shadows for inserted objects. The resulting model is efficient to fine-tune using only a small curated dataset, and substantially improves the photorealism of frames rendered from neural simulators.

\vspace{-2mm}
\section{Related Work}
\vspace{-2mm}
\label{sec:related}

\myparagraph{Image and Video Harmonization} aims to adjust the appearance of the inserted foreground objects so that they blend naturally with the background scene.   
Early approaches~\cite{tsai2017deep, cun2020improving, ling2021region, guo2021intrinsic} typically formulated harmonization as a frame-to-frame regression problem implemented using autoencoders. Ke \etal~\cite{ke2022harmonizer} introduced a white-box framework that learns interpretable filters aligned with human perception and preference. Xu~\etal~\cite{xu2023learning} proposed to first convert the input image into a linear color space, perform harmonization, and then transform it back to sRGB. More recently, with the advent of powerful generative models, diffusion-based approaches were explored, either through fine-tuning large pretrained diffusion models~\cite{ren2024relightful, chen2025zero} or via training-free formulations~\cite{lu2023tf}. Ljungbergh~\etal~\cite{ljungbergh2025r3d2} design a frame-wise enhancer trained on renders of reconstructed driving scenes with real dynamic actors replaced by recreated 3D assets. Unlike in our work, their target is only shadow casting and relighting, omitting artifacts that are common in neural reconstruction.

Extending the harmonization to video introduces the challenge of temporal coherence. Flow-based constraints~\cite{huang2019temporally} and local color mapping~\cite{lu2022deep} reduce flickering but still operate independently on each frame.  
More recent works process videos holistically: VHTT~\cite{guo2024video} introduces a Video Triplet Transformer modeling both global and dynamic temporal variations, while GenCompositor~\cite{yang2025gencompositor} employs a latent video diffusion model to harmonize inserted objects.  
However, existing models primarily address foreground harmonization and do not handle reconstruction artifacts or synthesize physically plausible shadows—which are essential for harmonizing frames rendered from neural reconstructed scenes used in simulation.

\myparagraph{Neural Reconstruction and Generative Priors.}
Neural reconstruction methods often fail in the sparse-view regime and struggle to generalize to novel viewpoints that are far from the training views, primarily due to the lack of strong data priors. To better constrain the optimization, early works introduced various geometric~\cite{kangle2021dsnerf, yariv2021volume, wang2021neus}, photometric~\cite{verbin2022refnerf}, and regularization-based priors~\cite{Niemeyer2021Regnerf}. 

More recent works have leveraged generative models to incorporate appearance priors and thus improve reconstruction quality. DiffusioNeRF~\cite{Wynn_2023_CVPR} leverages a learned scene geometry and appearance prior by training a diffusion model on RGB-D patches from synthetic data. GANeRF~\cite{roessle2023ganerf} leverages an adversarial formulation to refine a NeRF~\cite{mildenhall2020nerf} using a conditional generative network. Nerfbusters~\cite{Nerfbusters2023} incorporate the prior from a pretrained 3D diffusion model into the scene by using a density score distillation sampling loss~\cite{poole2022dreamfusion}. Several works rely on generating new views of the scene to use as supervision. ReconFusion~\cite{wu2023reconfusion} proposes a diffusion model that is conditioned on all input views through feature maps rendered by a PixelNeRF~\cite{pixelnerf} model. 3DGS-Enhancer~\cite{3dgsenhancer} employs a video diffusion model to generate novel views, Cat3D~\cite{gao2024cat3d} uses a multi-view latent diffusion model to generate novel views of the scene. 

In addition, DIFIX3D+~\cite{difix3d} proposes to use an image-to-image translation model to remove artifacts. Beyond usage to generate novel views to lift into 3D, the efficiency of the single-step model enables use at rendering time. Flowr~\cite{fischer2025flowr} proposes a multi-view flow matching model that learns to turn novel view renderings into clean images.

\begin{figure*}[th]
    \centering
    \vspace{-5mm}
    \includegraphics[width=0.95\linewidth]{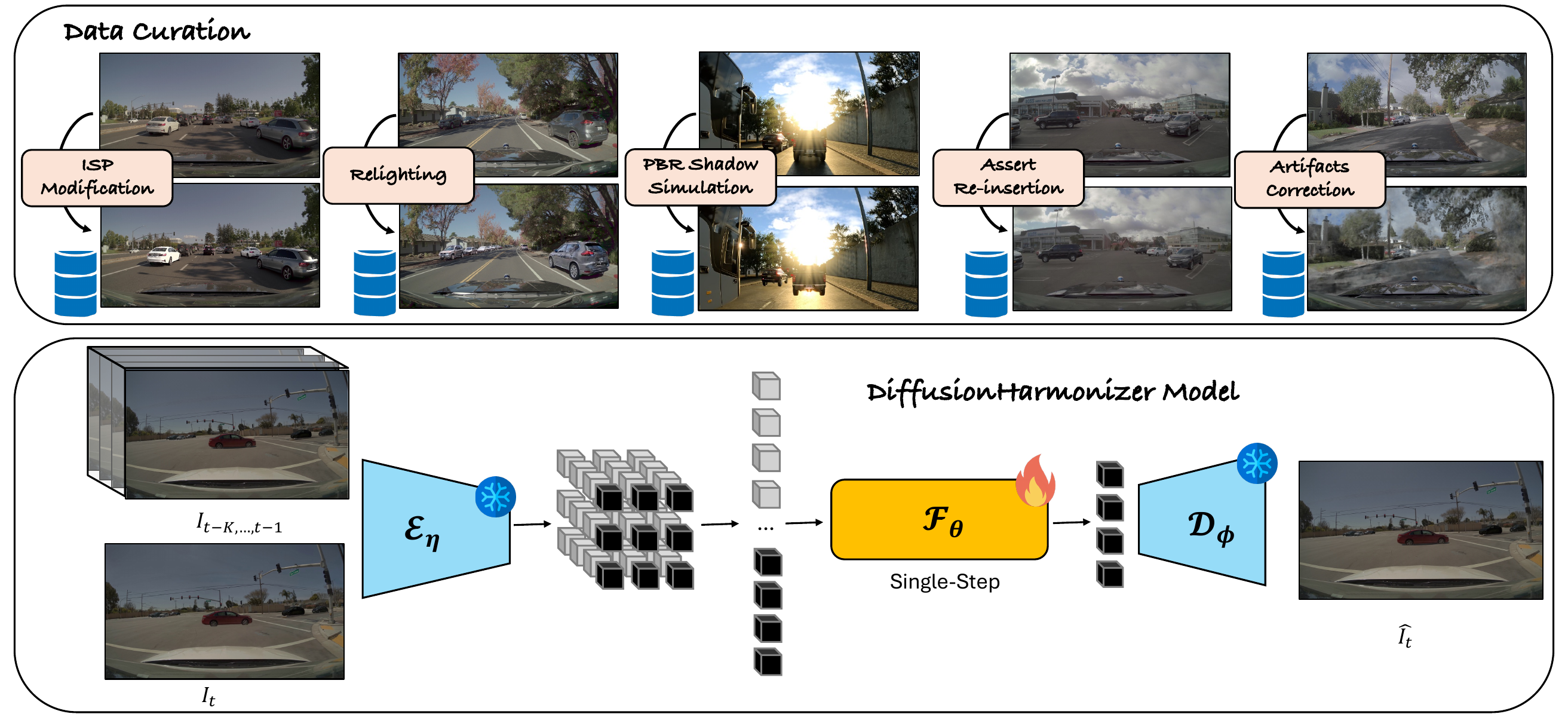}
    \vspace{-3mm}
    \caption{\textbf{Overview} of the data curation pipeline (\textbf{top}) and model architecture (\textbf{bottom}) of \fullname{}. We use a single-step temporally conditioned enhancement model, that is converted from a pretrained multi-step image diffusion model. To train it effectively, we develop a data curation pipeline to construct synthetic--real pairs emphasizing harmonization, artifact correction and lighting realism.}
    \vspace{-4mm}
    \label{fig:method_teaser}
\end{figure*}
\vspace{-2mm}

\section{Method}
\vspace{-1mm}
\label{sec:method}
Our goal is to develop an \emph{online harmonization model} that transforms artifact-prone rendered frames from reconstructed scenes into temporally consistent outputs with improved photorealism. 
This requires:  
(i) a lightweight architecture that enhances streaming frames while maintaining temporal stability, and
(ii) a curated paired dataset that supervises color harmonization, illumination consistency, and shadow synthesis.
We first describe the model architecture and training objectives, then our data curation pipeline.

\subsection{Online Frame-to-Frame Enhancer}

Pretrained diffusion models contain strong generative priors for image translation.  
We adapt an image-based diffusion model into a deterministic, single-step enhancer suitable for online simulation.

\myparagraph{Network Architecture.}
We formulate harmonization as an image-to-image translation task.  
Given a degraded frame $I_t$ at time $t$, the enhancer predicts an improved frame $\hat{I}_t$:
\vspace{-1mm}
\begin{equation}
\hat{I}_t = \mathcal{D}_\phi\!\left(\,\mathcal{F}_\theta\bigl(\mathcal{E}_\eta(I_t)\bigr)\right),
\end{equation}
where $\mathcal{E}_\eta$ and $\mathcal{D}_\phi$ denote the pretrained latent encoder and decoder, which are kept frozen, and $\mathcal{F}_\theta$ is the diffusion backbone.  
We fine-tune only $\mathcal{F}_\theta$ to adapt the pretrained diffusion model to the harmonization task.

In standard diffusion models, the backbone $\mathcal{F}_\theta$ is trained as a denoiser operating on stochastic noisy latents across many timesteps, with conditioning that encodes the noise level or diffusion time.
In contrast, we repurpose $\mathcal{F}_\theta$ as a \emph{deterministic} single-step enhancer. 
Specifically, we feed the clean latent $\mathcal{E}_\eta(I_t)$ directly into the network without injecting noise and fix the timestep and text-conditioning tokens to constant ``null'' values during both training and inference.
This produces a stable and deterministic mapping from input latent to enhanced latent and improves frame-wise structural consistency.

\myparagraph{Temporal Conditioning.}
For online simulation, it is crucial that per-frame enhancements remain temporally stable.
We therefore extend the backbone $\mathcal{F}_\theta$ to accept a short temporal context of previous frames.

Let $K$ denote the context length (we use $K=4$ in practice).
At time $t$, we encode the current degraded frame and up to $K$ previously enhanced frames:
\vspace{-1mm}
\begin{equation}
Z_t = \left[ \mathcal{E}_\eta(I_t), \; \mathcal{E}_\eta(\hat{I}_{t-1}), \dots, \mathcal{E}_\eta(\hat{I}_{t-K}) \right],
\end{equation}
and feed $Z_t$ into the backbone with temporal attention layers interleaved with spatial attention, following video diffusion architectures.
For the first few frames, when fewer than $K$ previous frames are available,
we condition only on the frames that exist.
This design allows the enhancer to use historical context when it is beneficial, while preserving frame-wise structure and preventing drift.

\subsection{Data Curation Strategy}\label{sec:data_curation}

Training our model requires paired data consisting of artifact-prone renderings and high-quality photorealistic images that capture diverse factors such as appearance harmonization, lighting realism, reconstruction robustness, and shadow consistency.  
Such supervision is scarce in real-world datasets and unavailable in existing public sources.  

To address this gap, we design a scalable data curation pipeline that synthesizes paired supervision data tailored for our enhancement objectives. 
The pipeline comprises five complementary components: 
\textit{Novel-view artifacts correction}, 
\textit{ISP modification}, 
\textit{Relighting}, 
\textit {Shadow simulation}, 
and \textit{Asset re-insertion}, 
with each targeting a specific visual factor. An overview is shown in \cref{fig:method_teaser}. 

\vspace{-1mm}
\myparagraph{Novel-View Artifacts Correction.}
To handle reconstruction artifacts in challenging novel view synthesis, we follow the data strategy of DIFIX3D+~\cite{difix3d}.
Specifically, we generate degraded renderings using four procedures: \emph{sparse reconstruction}, \emph{cycle reconstruction}, \emph{cross-referencing}, and deliberate \emph{model underfitting}. 
These operations produce frames with blurred details, missing regions, ghosting, and spurious geometry.
Every degraded frame is paired with its corresponding clean rendering, providing explicit supervision for correcting novel-view synthesis artifacts.

\vspace{-1mm}
\myparagraph{ISP Modification.}
Object captures from different devices often exhibit image signal processing (ISP) induced tone and color inconsistencies, leading to foreground–background appearance mismatches after composition.  
To simulate such mismatches, we implement a software ISP that re-renders captured images with randomized parameters (\eg, tone mapping, exposure, white balance, \etc).
Given an original capture $I_{\text{orig}}$,
we generate an alternative image $I_{\text{ISP}}$ using resampled ISP parameters.
Using a mask $M$ (obtained from SAM2~\cite{ravi2024sam}) that segments the foreground region,
we create a composite:
\vspace{-1mm}
\begin{equation}
I_{\text{mix}} = M \odot I_{\text{ISP}} + (1 - M) \odot I_{\text{orig}}.
\end{equation}
Here, $I_{\text{mix}}$ is used as the input and $I_{\text{orig}}$ as the target, enabling the model to learn foreground–background color and tone harmonization.

\vspace{-1mm}
\myparagraph{Relighting.}
To model illumination mismatches, we use an image relighting diffusion model~\cite{DiffusionRenderer} to regenerate selected regions under randomly sampled lighting conditions while preserving scene geometry and texture.
In practice, we relight only the foreground object, intentionally creating inputs where local illumination is inconsistent with the global scene lighting.  
These paired examples supervise the model to resolve lighting discrepancies and to synthesize spatially consistent illumination across the frame.

\vspace{-1.2mm}
\myparagraph{Physically Based Shadow Simulation.}
Accurate cast shadows are critical for realism but are difficult to annotate in real data.
To provide explicit supervision for shadow reasoning, we use a physically based renderer to synthesize cast shadows under controllable light configurations. 
We randomly vary the environment maps in synthetic scenes to modify the direction, softness, and intensity of the light source, and generate paired samples with and without shadows. These pairs provide precise pixel-level cues for the network to learn physically plausible shadow casting and attenuation. 

\vspace{-1.2mm}
\myparagraph{Asset Re-Insertion.}
While PBR-based shadow data offer precise supervision signals, they may not fully match real-world statistics.
To reduce the domain gap, we additionally leverage dynamic scene reconstruction. 
Specifically, we reconstruct the static background with 3DGUT~\cite{wu20253dgut}, extract dynamic foreground objects with an in-house solution, and then re-insert the foreground objects into the reconstructed background scene \textit{without} casting shadows. The resulting composites mimic the realistic object insertions but intentionally lack proper shadows and harmonization. 
Paired with the original sequences that contain correct shadows and coherent appearance, they offer rich supervision for learning realistic shadow synthesis and foreground–background harmonization.

Together, these components yield a diverse paired dataset that captures reconstruction artifacts, ISP-induced appearance mismatches, illumination inconsistencies, and missing shadows commonly seen in simulated renderings.
This curated data is a key enabler for training our online harmonizer under limited real-world supervision.

\subsection{Training}
\vspace{-1mm}
\myparagraph{Stabilizing Single-Step Training.}
Using a multi-step pretrained model in a single-step manner introduces a noise-trajectory mismatch:
the original model is trained to operate across a full diffusion trajectory, whereas at fine-tuning we apply it only once on a clean latent.  
Directly fine-tuning as a deterministic one-step model often leads to high-frequency artifacts such as checkerboard patterns.

To stabilize training, we introduce a multi-scale perceptual loss computed on randomly sampled squared patches of varying sizes. 
Given a predicted frame $\hat{I}_t$ and a ground-truth target frame $I^{\text{gt}}_t$, 
we sample square patches $\hat{P}_t^{(k)}$ and $P_{\text{gt}}^{(k)}$ of side length $k \in [128, 512]$ at random locations, and define the multi-scale perceptual loss as:
\vspace{-2mm}
\begin{equation}
\label{eq:stochastic-perc-loss}
\mathcal{L}_{\text{perc}} =
\mathbb{E}_{k}\!
\left[
  \sum_{l}
  \lambda_l \,
  \bigl\|
  \phi_l(\hat{P}_t^{(k)}) -
  \phi_l(P_{\text{gt}}^{(k)})
  \bigr\|_2^2
\right],
\end{equation}
where $\phi_l(\cdot)$ denotes features from the $l$-th layer of a VGG network and $\lambda_l$ are layer-wise weights.

Sampling multi-scale patches perturbs patch boundaries relative to the model's receptive field, emphasizing high-frequency inconsistencies and suppressing periodic aliasing.
Empirically, this loss significantly suppresses checkerboard artifacts arising from the noise-trajectory mismatch (see \cref{sec:ablation}).

\vspace{-1mm}
\myparagraph{Temporal Warping Loss.}
To further encourage temporal consistency, we incorporate a warping-based temporal loss.  
Given consecutive ground-truth frames $I^{\text{gt}}_{t-1}, I^{\text{gt}}_{t}$,
we estimate the optical flow $F_{t \rightarrow t-1}$ using RAFT~\cite{teed2020raft}.
We then warp the enhanced frame at time $t-1$ into time $t$ and enforce consistency in the visible pixels:
\vspace{-1mm}
\begin{equation}
\mathcal{L}_{\text{temp}} \!=\!
\frac{1}{|\Omega|}
\sum_{x \in \Omega}
\left[
\hat{I}_t(x) \!-\!
\mathrm{Warp}(\hat{I}_{t-1}, F_{t \rightarrow t-1})(x)
\right]^2,
\end{equation}
where $\Omega$ is the set of pixels for which the warping produces valid correspondences (i.e., not occluded or leaving the frame).
This loss is computationally tractable benefiting from our single-step formulation: we only need one forward pass per frame to obtain RGB outputs for supervision, avoiding the memory overhead of backpropagating through multi-step diffusion trajectories.

\vspace{-1mm}
\myparagraph{Mixed Temporal Training.}
Our curated dataset contains both short video sequences and standalone images.
The latter arise from components such as single-image relighting, for which high-quality temporal variants are difficult to synthesize. To leverage all available data and avoid overfitting to strong temporal cues, we train with mixed temporal and non-temporal batches.
The overall training objective is:
\vspace{-1mm}
\begin{equation}
\mathcal{L}_{\text{total}} =
\lambda_{l_2} \, \mathcal{L}_{l_2} +
\lambda_{perc} \, \mathcal{L}_{\text{perc}} +
\lambda_{temp} \, \mathcal{L}_{\text{temp}},
\end{equation} 
where $\mathcal{L}_{l_2}$ is a per-pixel loss between $\hat{I}_t$ and $I^{\text{gt}}_t$,
and $\lambda_{\text{temp}}=1$ for temporal batches and $0$ otherwise.

We first pretrain on paired image data to learn robust per-frame enhancement,
then alternate between temporal and non-temporal batches.
This schedule prevents the model from depending excessively on the nearby frames and improves robustness when temporal conditioning is weak, noisy, or partially unavailable.
\vspace{-1mm}

\section{Experiments}
\label{sec:experiments}

\vspace{1mm}
\begin{figure*}[t!]
  \centering
  \vspace{-3mm}
   \includegraphics[width=1.0\linewidth]{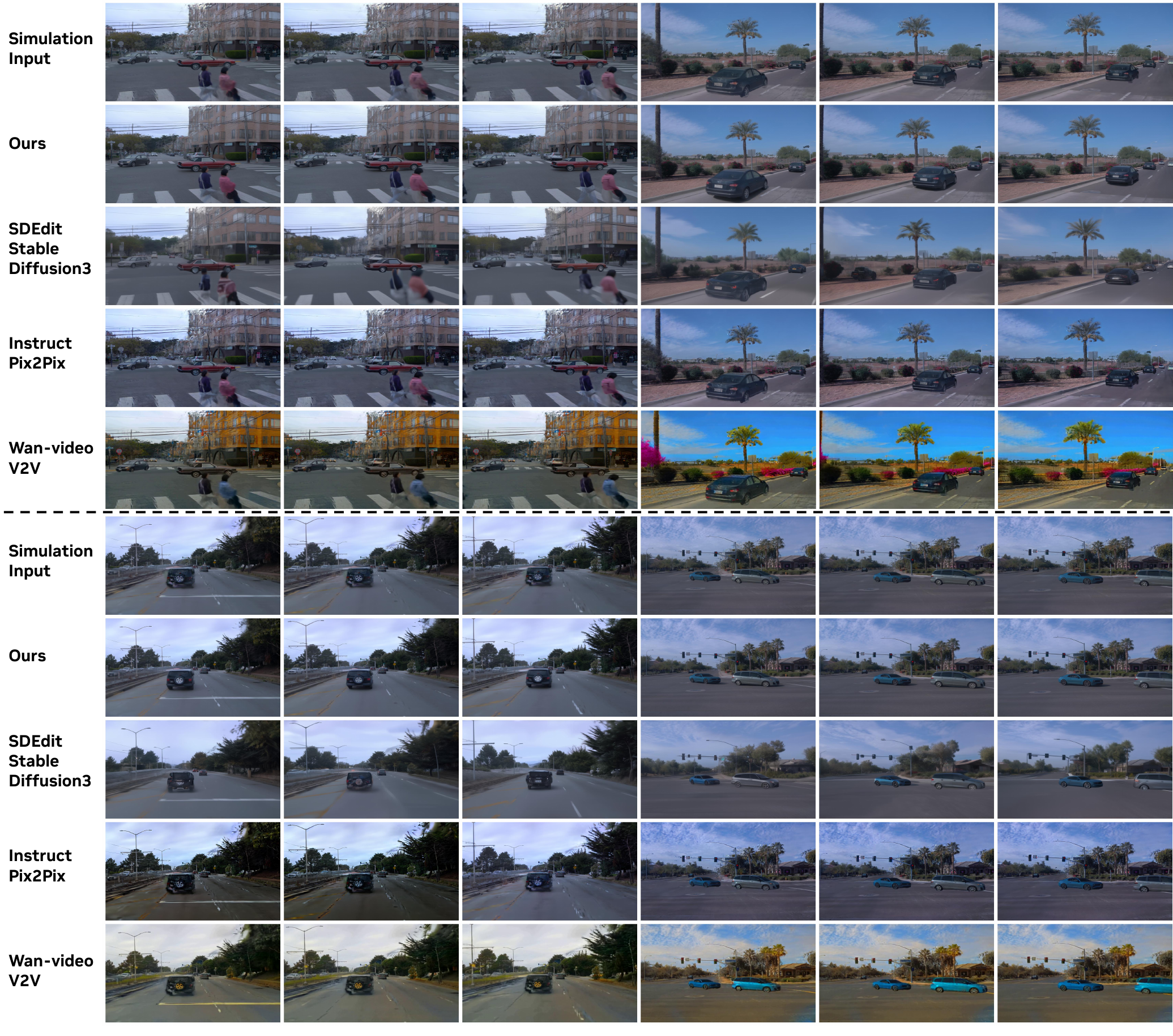}\\
\vspace{-4mm}
\caption{
\textbf{Comparison with Image and Video Editing Baselines on Out-of-Domain Testing Data.} Our method harmonizes color tone and synthesizes realistic lighting and shadows, while editing baselines often fail to produce physically plausible shadowing. Although both can reduce reconstruction artifacts, baselines tend to hallucinate inconsistent content and over-edit well-reconstructed regions, whereas our method preserves scene geometry and input structure. Moreover, image-editing baselines introduce frame-to-frame jitter, whereas our model maintains strong temporal coherence.}
\vspace{-4mm}
   \label{fig:qualitative}
   \vspace{0mm}
\end{figure*}

\begin{figure}[t]
    \centering
    \vspace{-0mm}
    \includegraphics[width=\linewidth]{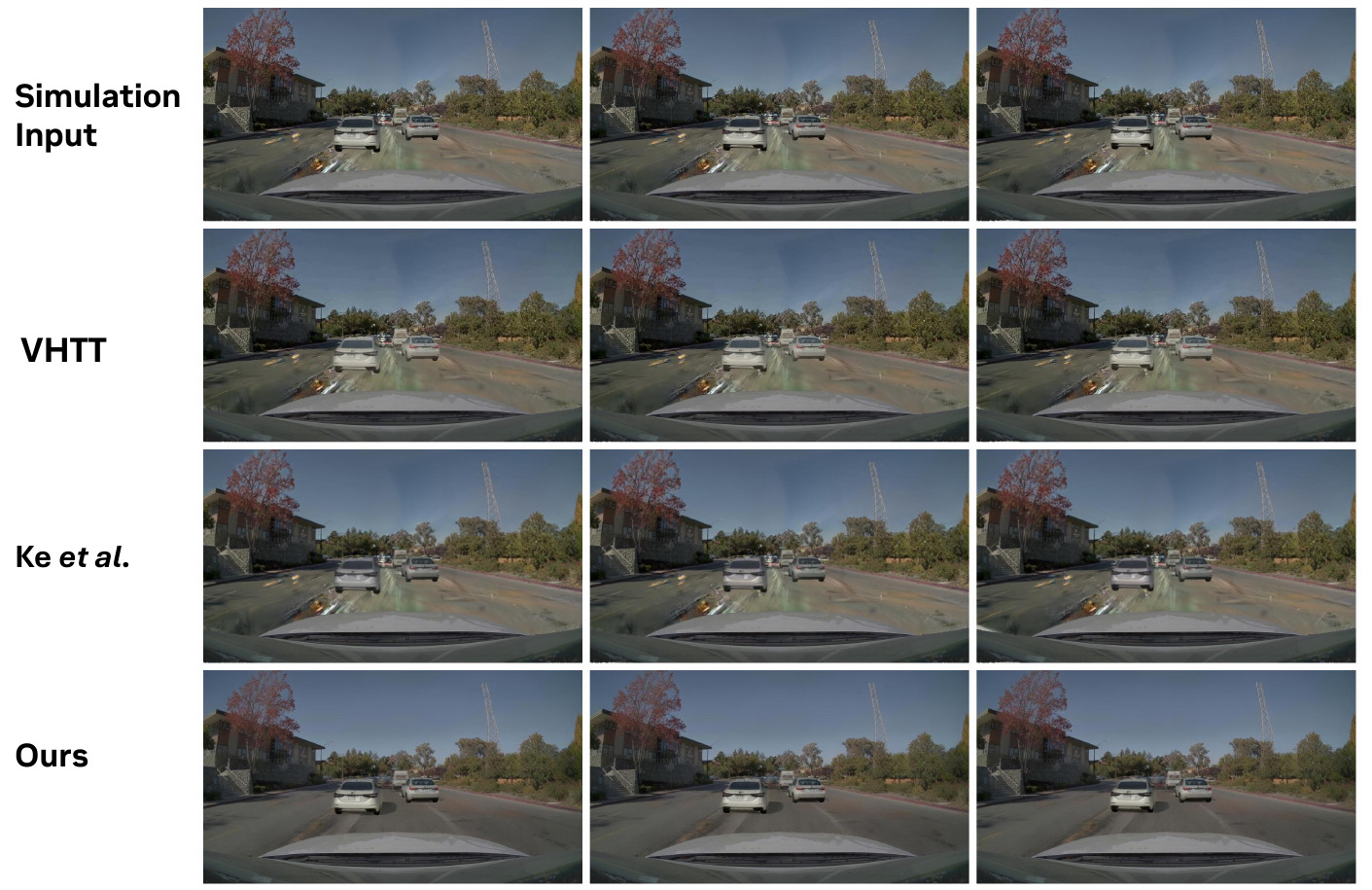}
    \vspace{-6mm}
    \caption{
    \textbf{Comparison with Harmonization Baselines Methods.} While both our method and harmonization baselines adjust foreground appearance, the baselines fail to synthesize realistic shadows, resulting in less coherent composites. }
    \label{fig:qualitative_harm}
    \vspace{-2mm}
\end{figure}
\vspace{-2mm}

We conduct extensive experiments across multiple datasets and evaluation settings to validate the effectiveness of our proposed framework. While tested on automotive scenarios, the method is domain-agnostic and readily applicable to other settings without modification.

\vspace{-1mm}
\myparagraph{Training Details.}
Our model is built upon the Cosmos 0.6B text-to-image diffusion model \cite{agarwal2025cosmos}, which contains 0.6B parameters in the diffusion backbone and 0.14B parameters in the VAE tokenizer. We freeze the VAE tokenizer during training and only fine-tune the diffusion backbone. The model is trained for 10k iterations for non-temporal pretraining and additional 4k for temporal training. We train the model at a resolution of $1024\times576$ with bf16 precision. We set $\lambda_{l2}=1$, $\lambda_{perc}=1$. 
Additional dataset statistics and visualizations are provided in the Supplement.

\vspace{-1mm}
\myparagraph{Baselines.} 
We compare our method against both general-purpose diffusion-based image and video editing models as well as task-specific harmonization approaches. 
\begin{itemize}
    \item \textit{General editing baselines:} We evaluate SDEdit~\cite{meng2021sdedit}, InstructPix2Pix~\cite{brooks2023instructpix2pix}, and Video-to-Video Editing (V2V) \cite{vace}. 
    For SDEdit, we replace the original model with the img2img variant of Stable Diffusion 3~\cite{esser2024scaling} to ensure state-of-the-art performance. For V2V, we use the VACE video-to-video editing model based on WAN 2.1~\cite{wan2025wan}. 
    
    \item \textit{Specialized video harmonization methods:}
    We compare our method with Ke~\etal~\cite{ke2022harmonizer} and VHTT~\cite{guo2024video}. Both methods were evaluated using official implementations and pre-trained checkpoints provided by the authors. 

\end{itemize}

\vspace{-1mm}
\myparagraph{Evaluation Protocol.}
We evaluate our model on simulated videos from three distinct settings. In all settings, we evaluate using 10-second video sequences rendered at 30FPS with resolution of 1024×576: 

\vspace{0.0em}
\begin{itemize}
    \item \textit{Novel Trajectory Simulation.} 
    We reconstruct both static scenes and dynamic objects (\eg, pedestrians and vehicles), then render novel views by laterally shifting the ego trajectory by 2\,m. Evaluation is performed on 13 holdout scenes from our internal driving dataset (in-domain test).

    \item \textit{Dynamic Object Insertion.} 
    We insert foreground objects (\eg, cars) into a pre-reconstructed scene and rendered in a novel trajectory shifted by 3 meters. Evaluation is performed on 68 scenes from the public Waymo driving dataset (out-of-domain test).  

    \item \textit{Holdout Datasets with Ground-Truth labels. } 
    We further evaluate our method on  holdout datasets for ISP modification, relighting, and PBR-based shadow simulation. These three datasets have accurate ground-truth labels, allowing the computation of GT-dependent evaluation metrics such as PSNR, SSIM, and LPIPS. 
\end{itemize}

\vspace{-2mm}
\myparagraph{Metrics.}
An ideal enhancement model should (1) improve perceptual realism, (2) preserve geometric and structural fidelity, and (3) maintain temporal consistency. We evaluate perceptual quality using {FID} and {FVD}, structural preservation using {DINO-Struct-Dist}, which measures feature-space similarity between input and output, and temporal smoothness using the temporal flickering score measured by {VBench++}. 
Because lighting, shadows, and general photorealism are difficult to quantify without ground-truth supervision, we additionally conduct the user study and, following recent practice \cite{kirstain2023pick, liu2024evalcrafter, lin2025controllable, bai2025qwen2}, employ a pretrained vision-language model (VLM) to assess overall quality. 
For holdout datasets with available ground-truth labels, we further calculate {PSNR}, {SSIM}, and {LPIPS} on the region of interest to assess the alignment with the ground-truth label.  We also report inference speed measured on a single H100 GPU.

\begin{table*}[t]
  \centering
  \vspace{-3mm}
  \resizebox{\linewidth}{!}{
  \begin{tabular}{@{}l|c|cccc|cccc@{}}
    \toprule
    \multicolumn{2}{c|}{} & \multicolumn{4}{c|}{Novel Trajectory Simulation (In-domain)} & \multicolumn{4}{c}{Object Insertion Simulation (Out-of-domain)} \\
    \cmidrule(lr){3-6} \cmidrule(l){7-10}
    Method & Inference Speed $\downarrow$ &
    FID$\downarrow$ & FVD$\downarrow$ & DINO Struct.$\uparrow$ & Temporal Cons.$\uparrow$ &
    FID$\downarrow$ & FVD$\downarrow$ & DINO Struct. $\uparrow$ & Temporal Cons. $\uparrow$ \\
    \midrule
    SDEdit (SD 3) & 398ms & 129.92 & 753.39 & 0.7075 & 0.9661 & 106.17 & 506.91 & 0.6345 & 0.9524 \\
    InstructPix2Pix & 555ms & 153.94 & 680.92  & 0.6339 & 0.9203 & 128.72 & 573.20 & 0.5793 & 0.9085 \\
    Wan-Video V2V & 2827ms & 134.98 & 506.86 & 0.8289 & \textbf{0.9828} & 104.42 & 474.96 & 0.8226 & \textbf{0.9675} \\
    \hline
    Our Model & \textbf{212ms} & \textbf{120.23} & \textbf{470.11} & \textbf{0.9215} & 0.9827 & \textbf{101.27} & \textbf{453.17} & 0.\textbf{9096} & 0.9670 \\
    \bottomrule
  \end{tabular}
  }
  \vspace{-2mm}
  \caption{\textbf{Quantitative Comparison on \emph{Novel Trajectory Simulation} (In-domain Test) and \emph{Object Insertion Simulation} (Out-of-domain Test) Datasets.} Our method outperforms all editing baselines in perceptual quality (lower FID/FVD) and preserves scene structure more faithfully (lower DINO-Struct-Dist). It also achieves strong temporal consistency (measured by VBench++ temporal flickering score), surpassing image-editing methods and matching video diffusion models, with only a marginal gap to WAN V2V.}
  \vspace{-5mm}
  \label{tab:quantitative}
\end{table*}

\begin{table}[t]
 \centering
  \resizebox{\columnwidth}{!}{
    \begin{tabular}{@{}c|*{3}{c}|*{3}{c}|*{3}{c}@{}}
      \toprule
      & \multicolumn{3}{c|}{Relighting Data (Images) } 
      & \multicolumn{3}{c|}{PBR Shadow Data (Videos)} 
      & \multicolumn{3}{c}{ISP Modification Data (Videos)} \\
      \cmidrule(lr){2-4} \cmidrule(lr){5-7} \cmidrule(l){8-10}
      Method &
      PSNR$\uparrow$ & SSIM$\uparrow$ & LPIPS$\downarrow$ &
      PSNR$\uparrow$ & SSIM$\uparrow$ & LPIPS$\downarrow$ &
      PSNR$\uparrow$ & SSIM$\uparrow$ & LPIPS$\downarrow$ \\
      \midrule
      SDEdit (SD 3) 
        & 15.17 & 0.9803 & 0.0098 
        & 15.65 & 0.9891 & 0.0095
        & 17.46 & 0.9850 & 0.0112 \\
      InstructPix2Pix 
        & 15.35 & 0.9812 & 0.0112 
        & 16.46 & 0.9905 & 0.0083
        & 16.71 & 0.9856 & 0.0105 \\
      Wan-Video V2V 
        & N/A & N/A & N/A 
        & 14.21 & 0.9867 & 0.098 
        & 15.81 & 0.9814 & 0.1335 \\
      \midrule
      Our Model 
        & \textbf{23.93} & \textbf{0.9938} & \textbf{0.0042} 
        & \textbf{26.31} & \textbf{0.9966} & \textbf{0.0028} 
        & \textbf{28.10} & \textbf{0.9974} & \textbf{0.0020} \\
      \bottomrule
    \end{tabular}
  }
  \vspace{-2mm}
  \caption{
  \textbf{Quantitative Results on \emph{Relighting}, \emph{PBR Shadow}, and \emph{ISP Modification} Holdout Sets.}
  Our method achieves substantially better PSNR, SSIM, and LPIPS, closely matching real-world references.} 
  \vspace{-4mm}
  \label{tab:quantitative_new}
\end{table}

\begin{table}
\centering
  \resizebox{\columnwidth}{!}{
    \begin{tabular}{@{}c|*{7}{c}@{}}
      \toprule
      \multirow{2}{*}{Method} & \multicolumn{7}{c}{ISP Modification Data} \\
      \cmidrule(l){2-8}
      & Speed & PSNR & SSIM & LPIPS & FID & FVD & Temporal Cons.\\
      \midrule
      VHTT     & 63ms & 20.96 & 0.9921 & 0.0049 & 46.23 & 168.83 & 0.9838\\
      Ke~\etal~\cite{ke2022harmonizer} & \textbf{10ms} & 25.98 & 0.9956 & 0.0037 & 61.51 & 170.16 & \textbf{0.9872}\\
      \midrule
      Ours     & 212ms & \textbf{28.58} & \textbf{0.9971} & \textbf{0.0021} & \textbf{42.03} & \textbf{158.27} & 0.9805 \\
      \bottomrule
    \end{tabular}
  }
  \vspace{-2mm}
  \caption{
  \textbf{Quantitative Comparison with Harmonization Baselines.}
  Evaluated on the harmonization subset (ISP Modification), our method outperforms all baselines. Inference speed reported at $1024\times576$ resolution for Ours and Ke~\etal~\cite{ke2022harmonizer}, and at $576\times320$ resolution for VHTT.} 
  \vspace{-3mm}
  \label{tab:quant_harm}
\end{table}

\vspace{-2 mm}

\subsection{Qualitative Comparison}
\cref{fig:qualitative} presents a qualitative comparison between our method and state-of-the-art image and video editing baselines, on out-of-domain testing dataset. Our approach effectively harmonizes color tone and synthesizes realistic lighting and shadows, whereas baseline editing models often fail to generate physically plausible shadowing for inserted objects. Although both our method and existing editing models can mitigate reconstruction artifacts such as missing details or blurry regions, baseline outputs frequently hallucinate content inconsistent with the underlying input scene structure and also modify well-reconstructed regions that should remain unchanged. In contrast, our method faithfully preserves scene geometry and underlying input structures. Furthermore, while image-editing baselines can maintain coarse scene consistency under a fixed random seed, they typically exhibit frame-to-frame variation in fine details, resulting in temporal jitter. 
Our model achieves substantially better temporal coherence across adjacent frames.

We additionally compare against specialized harmonization methods in \cref{fig:qualitative_harm}. 
While both our method and baseline harmonization approaches adjust foreground color tone to match the background, these baselines fail to synthesize 
realistic shadows, leading to less realistic composites. Moreover, harmonization methods do not correct reconstruction artifacts by their design, limiting their applicability in neural simulation pipelines. Finally, we present more examples and also further visualize our model's predictions on holdout datasets with ground-truth label in the Supplementary materials. 

\subsection{Quantitative Evaluation}
We split our quantitative evaluation into two segments: (1) image and video editing baseline comparison, and (2) video harmonization baseline comparison.

\vspace{-1mm}
\myparagraph{Image and Video Editing Baselines.}
As shown in \cref{tab:quantitative}, our method consistently outperforms all editing baselines in perceptual quality, achieving better FID and FVD scores. Compared to the baselines, our model also preserves scene structure more faithfully, as reflected by markedly lower DINO-Struct-Dist scores. For temporal consistency, our approach significantly outperforms image-editing baselines and achieves performance comparable to video diffusion models, evidenced by the marginal difference in the VBench temporal score relative to WAN V2V. On holdout datasets with ground-truth label (\cref{tab:quantitative_new}), our predictions exhibit much closer alignment, achieving large improvements over all baselines in PSNR, SSIM, and LPIPS. Moreover, our method is at least 1.8$\times$ faster than image-editing baselines and 10$\times$ faster than video-editing baselines, enabling online deployment.

\vspace{-1mm}
\myparagraph{Video Harmonization Baselines.} 
We compare against harmonization baselines on the ISP modification dataset (where segmentation masks are available, required by the harmonization baselines), and report the results in \cref{tab:quant_harm}. Our method consistently outperforms VHTT \cite{guo2024video} and Ke~\etal~\cite{ke2022harmonizer} across all image-quality metrics.

\vspace{-1mm}
\myparagraph{User Study.} We conduct a user study to assess the overall quality. For each comparison, human evaluators are shown: (1) the input image as reference, and (2) a pair of predictions---our output and a baseline output---and asked to select the better result. The order of the predictions is randomized to avoid systematic bias. We recruit 45 evaluators (15 per baseline), each completing 50 pairwise comparisons. We report more configuration details in the supplementary materials. 
In \cref{tab:userstudy}, we report the mean preference rate and standard deviation across evaluators. As shown in the results, human evaluators consistently prefer our method over all baselines.

\begin{table}
  \centering
  \resizebox{\columnwidth}{!}{
  \begin{tabular}{@{}l|ccc@{}}
    \toprule
    \multirow{3.75}{*}{Evaluation} & \multicolumn{3}{c}{Baselines} \\
    \cmidrule(l){2-4}
     & SDEdit (SD 3) & InstructPix2Pix & Wan-Video V2V \\
    \midrule
    Human Eval. (\%) & $84.28\% \pm 10.92\%$ & $90.10\% \pm 14.54\%$  & $90.11\% \pm 14.13\%$ \\
    VLM Eval. (\%)  & 79.18\% & 75.41\% & 88.57\% \\
    \bottomrule
  \end{tabular}
  }
  \vspace{-2.5mm}
  \caption{
  \textbf{User Study.} Both human participants and VLM evaluators were asked to compare our results against each individual baseline and select the better one. We report the percentage of samples where our method is preferred over the baselines. A preference rate above 50\% indicates that our method is preferred.}
  \vspace{-2mm}
  \label{tab:userstudy}
\end{table}
\vspace{-1mm}

\subsection{Ablation Study}
\label{sec:ablation}
We analyze the effectiveness of our model design and data curation strategy through extensive ablation studies.

\begin{figure}[t]
    \centering
    \vspace{-0mm}
    \includegraphics[width=\linewidth]{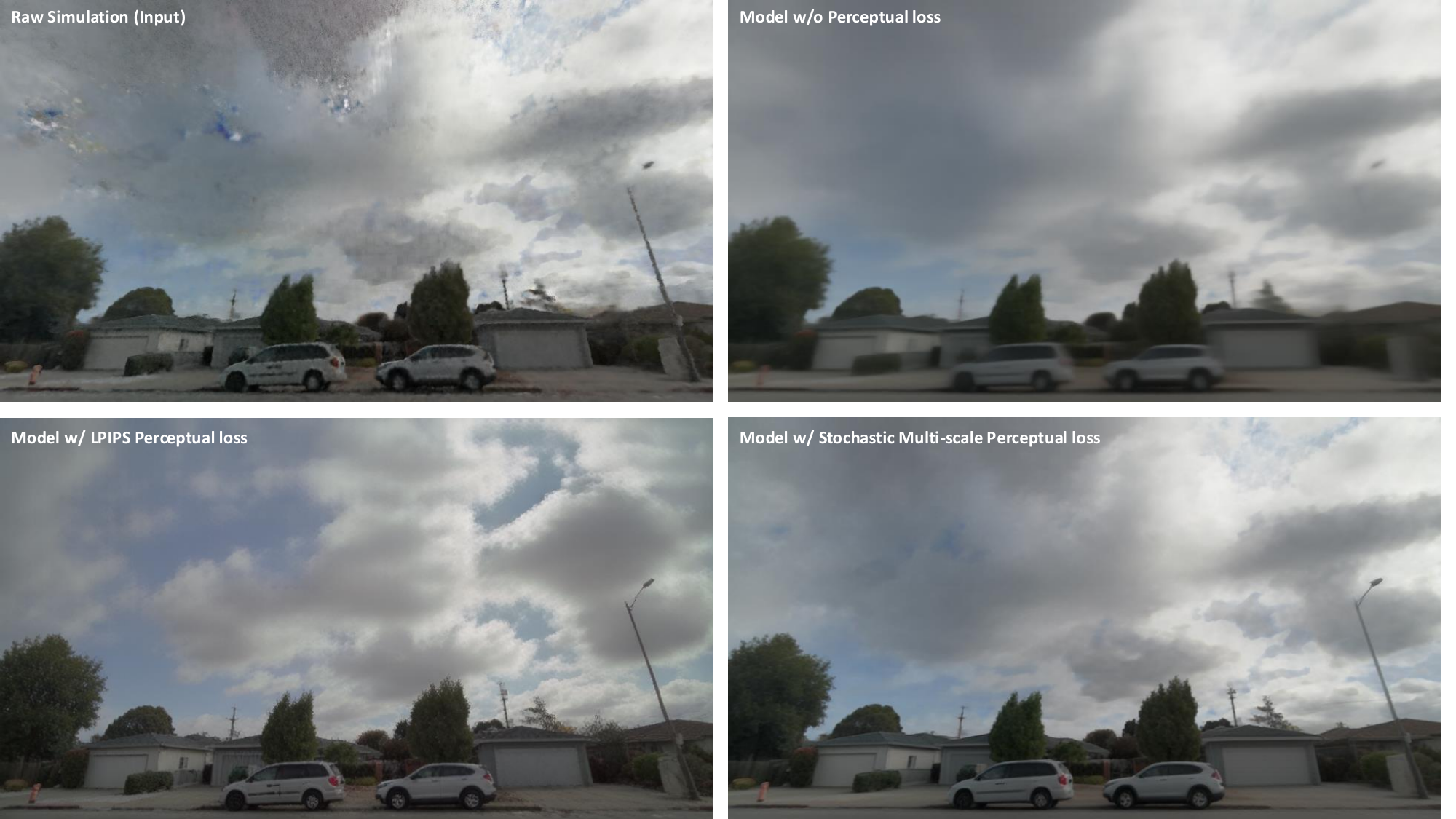}
    \vspace{-6mm}
    \caption{\textbf{Ablation on Loss Design.}  Removing perceptual supervision leads to oversmoothed outputs, while using a conventional LPIPS loss produces high-frequency artifacts. Our multi-scale formulation mitigates these artifacts and yields perceptually better results.}
    \label{fig:percept}
    \vspace{-3mm}
\end{figure}

\vspace{-1mm}
\myparagraph{Multi-Scale Perceptual Loss.}
We assess the impact of our multi-scale perceptual loss introduced in \cref{eq:stochastic-perc-loss}. As shown in \cref{fig:percept}, removing perceptual supervision leads to oversmoothed outputs, while using a conventional LPIPS loss produces high-frequency artifacts. Our multi-scale formulation mitigates these artifacts and yields perceptually better results. 

\vspace{-1mm}
\myparagraph{Temporal Components.}
We evaluate the effect of temporal conditioning and temporal loss design by comparing models trained without the temporal modules and temporal loss~$L_{\text{temp}}$ and show results in Table \ref{tab:abl_temporal}. The inclusion of temporal components improves temporal consistency metrics. Qualitatively, this manifests as reduced flickering and smoother transitions across frames. 

\begin{table}
  \centering
  \vspace{-2mm}
  \resizebox{\columnwidth}{!}{
  \begin{tabular}{@{}l|cc@{}}
    \toprule
    \multirow{2}{*}{Model Variant} & \multicolumn{2}{c}{Temporal Consistency Score} \\
    \cmidrule(l){2-3}
     & In-domain Test & Out-of-domain Test \\
    \midrule
    Our Model & 0.9827 & 0.9670 \\
    Our Model w/o Temporal Loss & 0.9806 & 0.9618 \\
    Our Model w/o Temporal Modules & 0.9714 & 0.9502 \\
    \bottomrule
  \end{tabular}
  }
  \vspace{-2.5mm}
    \caption{
    \textbf{Ablation on Temporal Components.} Adding temporal loss and temporal modules effectively improves temporal consistency. 
    } 
  \vspace{-5mm}
  \label{tab:abl_temporal}
\end{table}

\begin{table}[h!]
    \centering
     \resizebox{\columnwidth}{!}{
    \begin{tabular}{lcccccc}
    \toprule
    \textbf{Metric} & \textbf{Full} & \textbf{w/o ISP} & \textbf{w/o } & \textbf{w/o Shadow} & \textbf{w/o Asset} & \textbf{w/o Artifacts} \\
     & \textbf{Model} & \textbf{Modif.} & \textbf{Relight.} & \textbf{Simul.} & \textbf{Re-ins.} & \textbf{Correc.} \\
    \midrule
    FID $\downarrow$ & \textbf{101.27} & 104.63 & 102.15	& 104.28 & 103.42 & 105.29   \\
    FVD $\downarrow$ & \textbf{453.17} & 465.7 & 457.92 & 462.31 & 459.34 & 476.82  \\
    \bottomrule
    \end{tabular}
    }
    \vspace{-2.5mm}
    \caption{\textbf{Quantitative ablation on curated data sources.} Model trained on all data sources provides the best performance.}
    \label{tab:ablation_data}
   \vspace{-6mm}
\end{table}

\vspace{-2mm}
\myparagraph{Data Curation Strategy.} 
We further ablate the contribution of different data sources from our curation strategy. \cref{tab:ablation_data} quantifies individual contributions of separate data curation streams to the overall model performance confirming that all curated data sources contribute complementary supervision signals critical for generalization. Further qualitative results can be found in \cref{fig:data_abl} in the Supplementary materials.

\vspace{-2mm}

\section{Conclusion}
\vspace{-1mm}
We propose \fullname{}, an online generative enhancement framework that transforms artifact-prone neural-rendered frames into photorealistic and temporally coherent simulations. By converting a pretrained image diffusion model into a single-step temporally conditioned enhancer, and by introducing a comprehensive data-curation pipeline together with a tailored training objective, our method effectively addresses reconstruction artifacts, appearance inconsistencies, illumination mismatches, and missing shadows. Extensive experiments demonstrate substantial gains over both image/video editing and harmonization baselines in perceptual realism, structural fidelity, and temporal stability, while operating efficiently on a single GPU. We believe \fullname{} offers a practical and scalable solution for high-fidelity simulation in autonomous driving and robotics, and opens new avenues for integrating generative priors into real-time simulation pipelines.
\label{sec:conclusion}
\clearpage
{
    \small
    \bibliographystyle{ieeenat_fullname}
    \bibliography{main}
}
\clearpage
\setcounter{page}{1}
\maketitlesupplementary

We provide preliminary background information in \cref{sec:preliminary}, additional experimental details in \cref{sec:exp_details}, and further qualitative results in \cref{sec:additional_qualitative}. We provide additional video comparisons in the attached video files in the supplementary materials.

\section{Background Information}
\label{sec:preliminary}

\noindent\paragraph{Diffusion Models} are generative models that learn to transform samples from one source distribution, typically Gaussian noise, to the target data distribution $p_{\text{data}}(\rvx)$ \cite{ddpm, sohldickstein, scorebased}. They consist of two main processes: a \emph{forward process} where noise $\rvepsilon \sim \gN(\mathbf{0}, \mI)$ is progressively added to samples $\rvx$ from the data distribution, and a \emph{reverse process} which iteratively removes noise from a sample from the source distribution to obtain a sample of the data distribution. The intermediate latent variables can be expressed as $\rvx_\tau= \alpha_\tau\rvx + \sigma_\tau\rvepsilon$ where $\alpha_\tau$ and $\sigma_\tau$ are determined by the selected noise schedule and the timestep $\tau$ represents how far along the forward process we are (higher $\tau$ indicates more noise). The reverse process is achieved by a denoising model $\mathbf{F}_\theta$ with learnable parameters $\theta$. These are optimized by learning to predict the added noise; concretely they minimize the following objective: 
\begin{align}
\E_{\rvx \sim p_{\text{data}}, \tau \sim p_{\tau}, \rvepsilon \sim \gN(\mathbf{0}, \mI)} \left[ w(\tau) \Vert \rvepsilon - \mathbf{F}_\theta(\rvx_\tau; \rvc, \tau) \Vert_2^2 \right],
\label{eq:diffusionobjective}
\end{align}
where $\vc$ represents the condition (\ie, text prompt or reference image) used to control the generation, $w(\tau)$ is a time-dependent weighting function. $p(\tau)$ is a chosen timestep distribution, typically a uniform distribution over a set of integers, i.e. $p_\tau \sim \mathcal{U}(0,1000)$ \cite{ddpm}. Latent Diffusion Models (LDMs) \cite{ldm} greatly improve computational and memory efficiency by operating on a lower dimensional latent space. The dimensionality reduction is achieved by an encoder-decoder model, where the encoder $\mathcal{E}$ maps the data samples into a latent space $\mathbb{Z}$ and the decoder $\mathcal{D}$ achieves the inverse operation: $\mathcal{D}\left(\mathcal{E}(\rvx)\right)\approx \rvx$. LDMs \cite{ldm} effectively treat $\mathbb{Z}$ as the data set, therefore $\rvx$ can be replaced by $\rvz:=\mathcal{E}(\rvx)$ in all equations above.   

\section{Additional Experiment Details}     
\label{sec:exp_details}

\begin{figure*}[t!]
  \centering
  \vspace{-3mm}
\includegraphics[width=1.\linewidth]{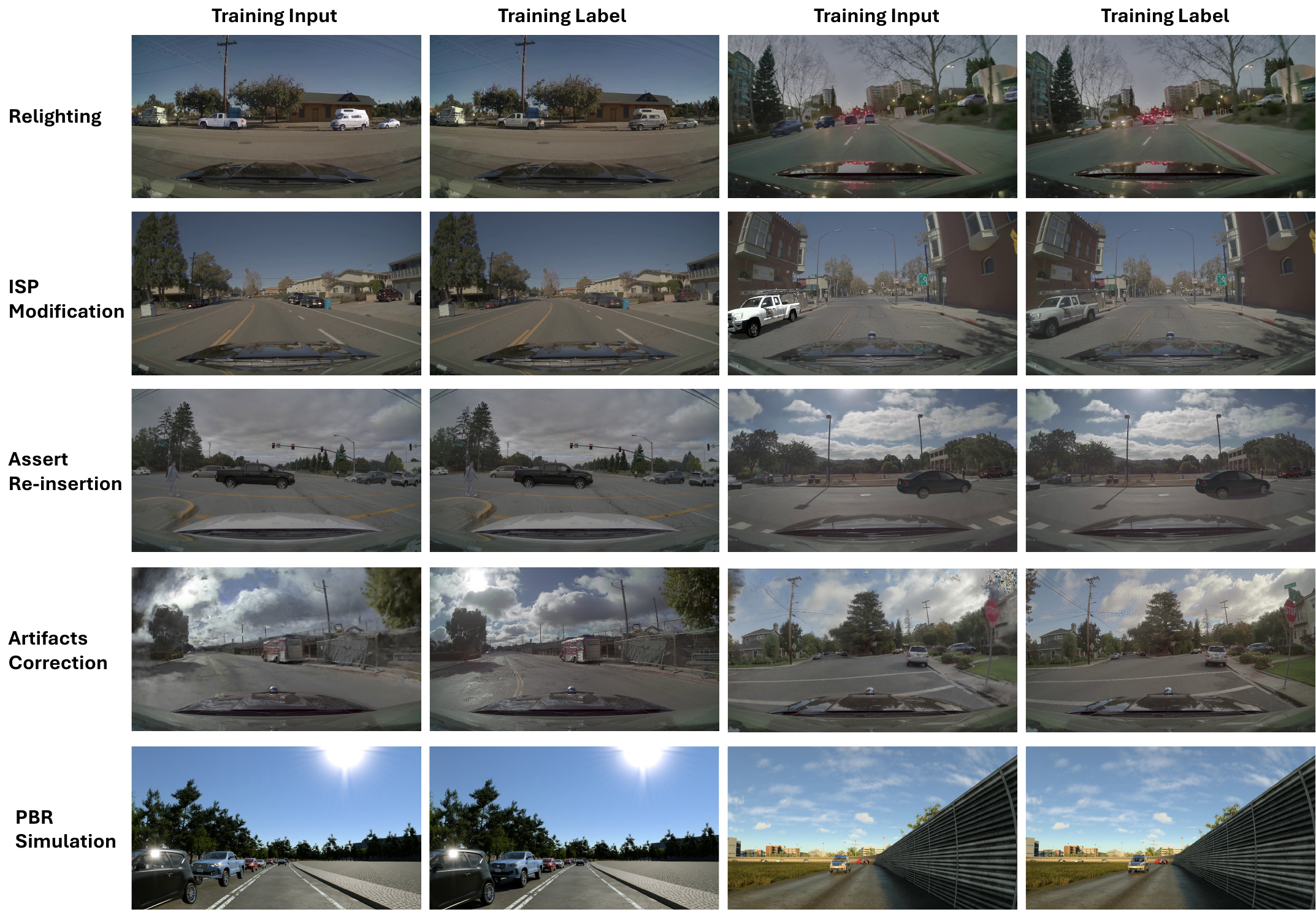}\\
\vspace{-4mm}
\caption{\footnotesize
\textbf{Visualization of Curated Training Dataset.} We show representative paired samples from our curated training data: Relighting, ISP Modification, Asset Re-insertion, Artifacts Correction, and PBR Simulation.}
\vspace{-4mm}
   \label{fig:data_visual}
   \vspace{0mm}
\end{figure*}

\myparagraph{Training Dataset Details:} We visualize in \cref{fig:data_visual} representative paired samples from the five components of our data-curation pipeline: Relighting, ISP Modification, Asset Re-insertion, Artifacts Correction, and PBR Simulation. Each group shows the training input and its corresponding label, illustrating the types of appearance changes, lighting variations, reconstruction degradations, and shadow differences used to supervise DiffusionHarmonizer. The total training set includes 46K frames for Relighting, 88K frames for ISP Modification, 21K frames for Asset Re-insertion, 118K frames for Artifacts Correction, and 77K frames for PBR Simulation. These pairs collectively provide diverse and complementary signals for harmonization, artifact correction, and lighting consistency. 

\myparagraph{General Image/Video Editing Baselines:} SDEdit and InstructPix2Pix require user-provided prompts and selections of hyperparameters (\eg., the magnitude of injected noise). To minimize human-induced prompting bias, we automatically generated all image-editing prompts using ChatGPT-5 and adopted the default hyperparameters provided by the official HuggingFace implementations. The generated prompt we use is:
\begin{quote}
\textit{"Translate this image into a high-quality camera frame captured by an autonomous vehicle, adding realistic shadows while correcting artifacts and harmonizing lighting and appearance between the foreground and background."}
\end{quote}
The corresponding negative\_prompt is:
\begin{quote}
\textit{"blurry, low quality, distorted, artifacts"}
\end{quote}

For video-to-video editing, we employ the VACE model built on WAN 2.1~\cite{wan2025wan}. Among the supported tasks, we select the grayscale-to-RGB translation setting as it is the closest analogue to our problem. During inference, we supply the model with (i) a prompt describing the video content and (ii) the grayscale frames as input, and request the model to synthesize the corresponding RGB sequence. All video prompts are generated using Qwen3-VL.

\begin{figure*}[t!]
  \centering
  \vspace{-3mm}
   \includegraphics[width=0.8\linewidth]{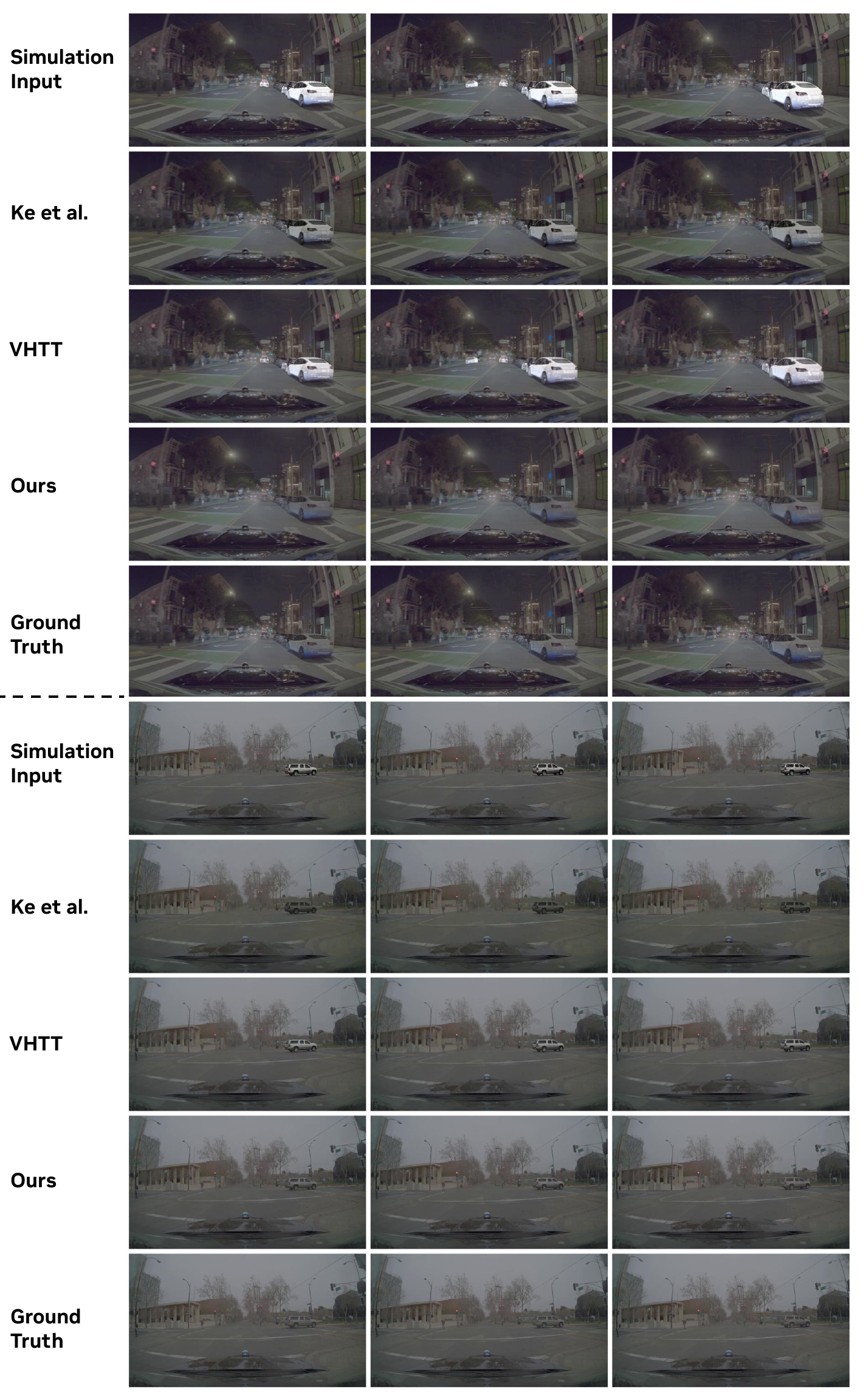}\\
\vspace{-4mm}
\caption{\footnotesize
\textbf{Additional Comparison with Video Harmonization Baselines on ISP modification held out test set.}}
\vspace{-4mm}
   \label{fig:qualitative_supp_videoharm2}
   \vspace{0mm}
\end{figure*}

\myparagraph{Specialized Video Harmonization Methods:} VHTT implements a design based on joint processing of video frames and stacked short- and long-term context transformer modules, which imposes limits on the resolution and length of the processed videos. VHTT video evaluations were thus carried out on batches of 20 frames at $576\times320$ resolution. All results were upsampled to $1024\times576$ resolution for quantitative evaluation. As VHTT~\cite{guo2024video} and Ke et al.~\cite{ke2022harmonizer} only alter pixels of inserted foreground objects and require input segmentation masks, the metrics reported in \cref{tab:quant_harm} were computed on pixels within the foreground regions only. As \cref{tab:quant_harm} shows, our method significantly outperforms both baselines in PSNR, SSIM, LPIPS, FID, and FVD, highlighting its power as an image and video harmonizer. Note that both methods assume the presence of a single foreground object, which may lead to degraded performance when applied to images with multiple inserted objects. This requirement is highly impractical in real-world scenarios, where one often deals with complex, multi-object scenes without access to instance segmentation masks. In contrast, our method operates in a mask-free setup, automatically identifying and harmonizing dissonant regions while having access to the full image, allowing it to enhance the entire scene. This is also reflected in inference speed: while our method generates a full image, Ke et al.~\cite{ke2022harmonizer} only predicts parameters of six image filters that are then sequentially applied to the foreground area. Figures~\ref{fig:qualitative_supp_videoharm} and \ref{fig:qualitative_supp_videoharm2} show additional qualitative comparisons. Notice how our method provides superior results even in a scene with only a single inserted object in \cref{fig:qualitative_supp_videoharm2} (bottom).

\begin{figure*}[t!]
  \centering
  \vspace{-3mm}
   \includegraphics[width=1.\linewidth]{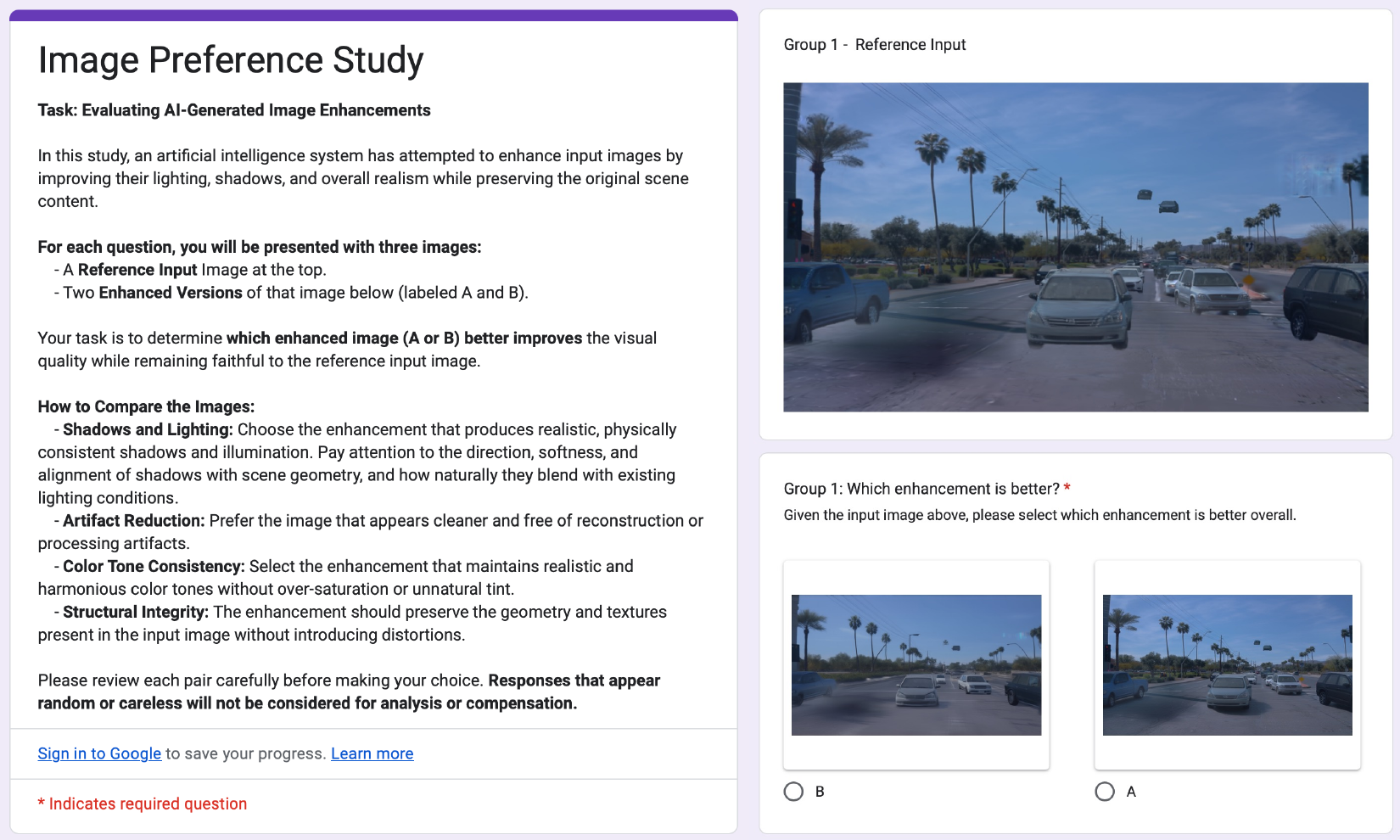}\\
\vspace{-2mm}
\caption{\footnotesize
\textbf{User Study Interface.} We show our study instructions and interface. Evaluators are shown the input image and two predictions (ours and a baseline) and asked to select the more realistic result, with prediction order randomized to avoid bias.}
\vspace{-4mm}
   \label{fig:user_study}
   \vspace{0mm}
\end{figure*}

\myparagraph{User Study \& VLM-Based Evaluation:} We visualize our study instructions and the user-study interface in \cref{fig:user_study}. During the study, human evaluators are shown (i) the input image as a reference and (ii) two predictions—our output and a baseline—and are asked to choose the result they perceive as more realistic. To mitigate ordering bias, the left–right placement of the two predictions is randomized for every comparison. Participants are recruited through the Prolific platform, which provides access to a diverse, globally distributed pool of evaluators. In total, we gather responses from 45 participants (15 per baseline), with each participant completing 50 pairwise comparisons.

To complement the human study and provide a scalable, automated comparison, we additionally evaluate model preference using a vision–language model (VLM), specifically the LLaVA-1.5-7b model. Following the same protocol as the user study, we present the VLM with the input image as a reference along with two predictions—our output and a baseline—with their order randomly shuffled. The model is asked to choose the result that better preserves realism and consistency relative to the reference. To further reduce prompt-design bias, all evaluation prompts are automatically generated using ChatGPT-5. Results in \cref{tab:userstudy} show a high level of agreement between the VLM-based judgments and human preferences, reinforcing the superior quality of our method.

\section{Additional Qualitative Results}
\label{sec:additional_qualitative}

\subsection{Qualitative Ablation of Curated Data Sources}
\begin{figure}[t]
    \centering
    \includegraphics[width=\linewidth]{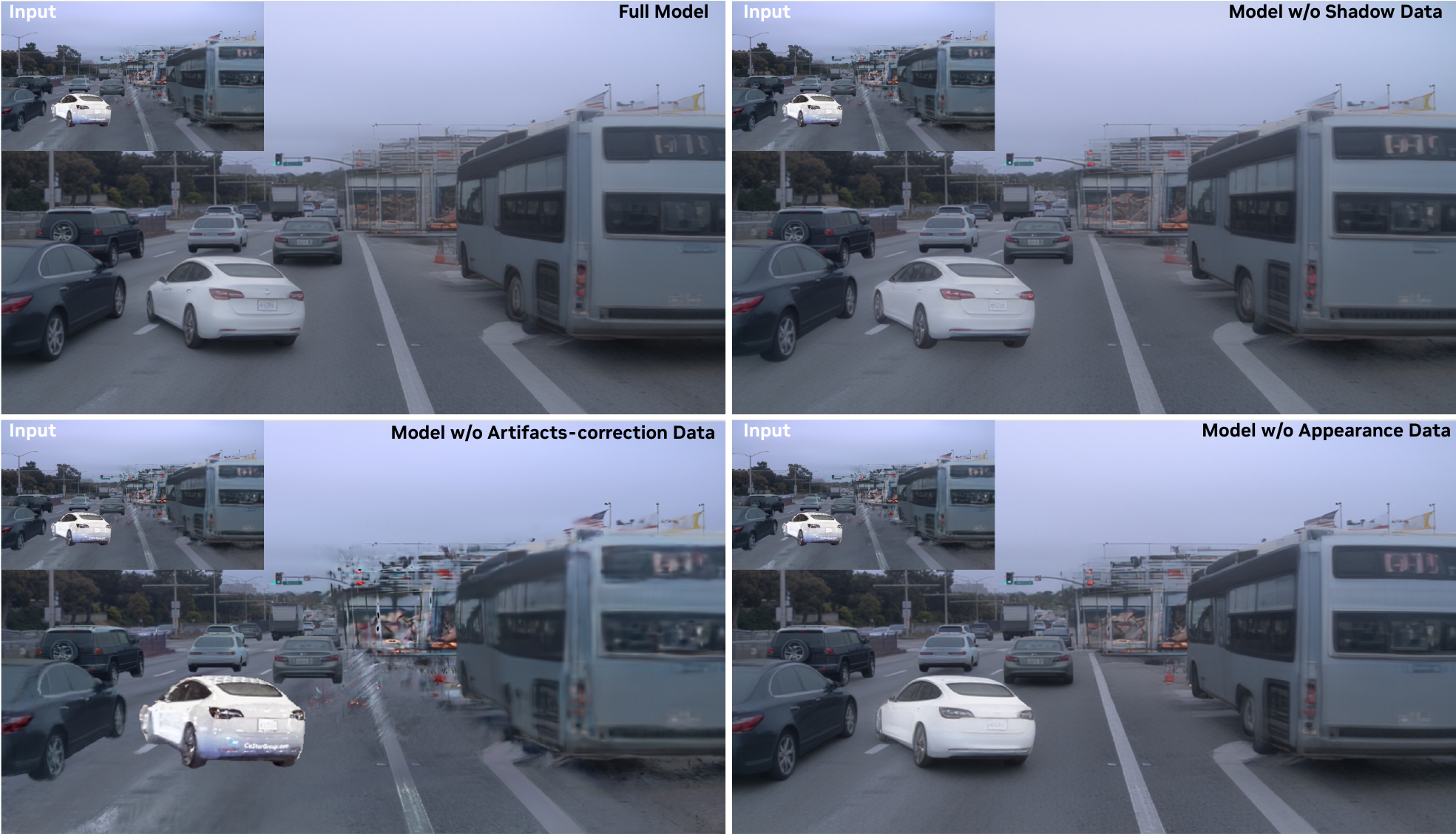}
    \vspace{-6mm}
    \caption{
    \textbf{Ablation on Curated Data Sources.} Removing any curated data source degrades performance: without artifact-correction data the model fails to fix reconstruction errors; without shadow data it cannot synthesize plausible shadows; and without appearance data it produces color-tone inconsistencies. Each data source provides complementary and essential supervision.}
    \label{fig:data_abl}
    \vspace{-3mm}
\end{figure}

\cref{fig:data_abl} highlights the qualitative impacts of ablating individual training data subsets: excluding Artifacts-Correction data prevents the model from correcting reconstruction errors; removing Shadow Data (PBR Shadow and Asset Re-Insertion) prevents synthesizing physically plausible shadows; and removing the appearance dataset (ISP Modification and Relighting data) results in color-tone discontinuities between foreground and background. This result together with the quantitative analysis in \cref{tab:ablation_data} underscores the importance of all components for the model performance.

\subsection{Additional Qualitative Comparison with Baselines}

\begin{figure*}[t!]
  \centering
  \vspace{-3mm}
  \includegraphics[width=1.\linewidth]{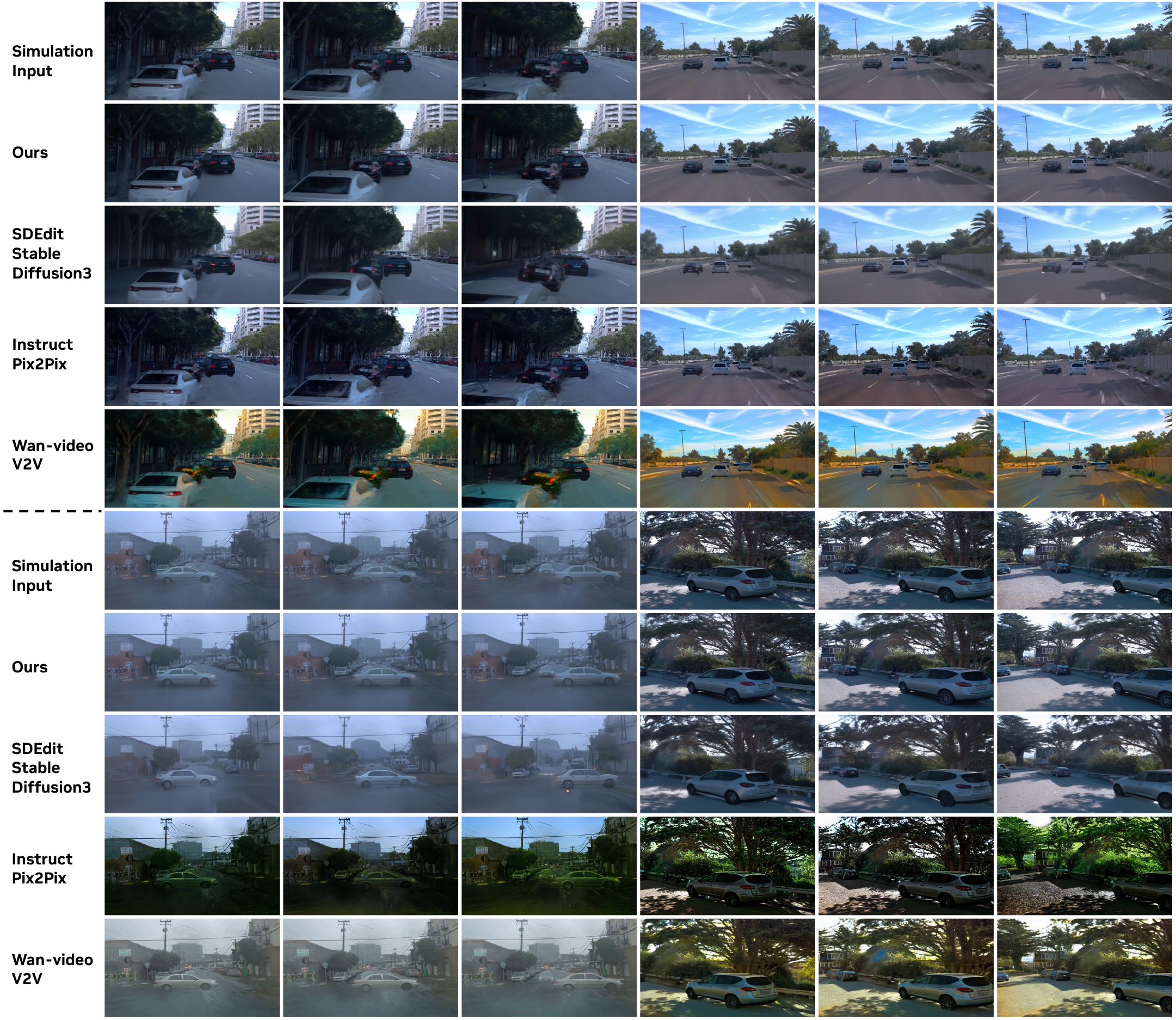}
  \vspace{-4mm}
  \caption{\footnotesize
  \textbf{Additional Comparison with Image and Video Editing Baselines on Out-of-Domain Testing Data (Part 1 of 2).}
  Our method harmonizes color tone and synthesizes realistic lighting and shadows, while editing baselines often fail to produce physically plausible shadowing. Although both can reduce reconstruction artifacts, baselines tend to hallucinate inconsistent content and over-edit well-reconstructed regions, whereas our method preserves scene geometry and input structure. Moreover, image-editing baselines introduce frame-to-frame jitter, whereas our model maintains strong temporal coherence.
  }
  \label{fig:qualitative_supp_part1}
  \vspace{-4mm}
\end{figure*}

\begin{figure*}[t!]
  \centering
  \vspace{-3mm}
  \includegraphics[width=1.\linewidth]{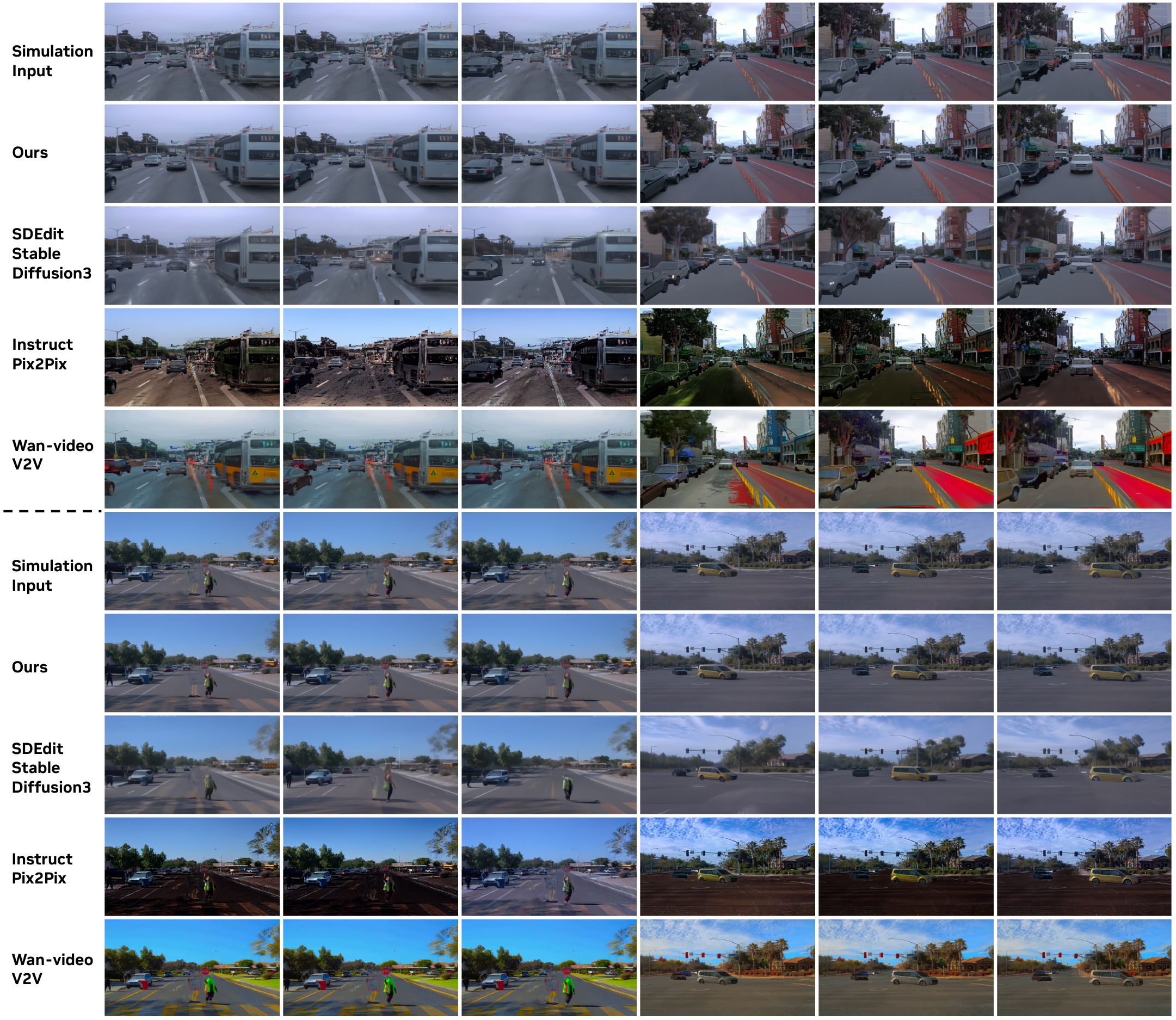}
  \vspace{-4mm}
  \caption{\footnotesize
  \textbf{Additional Comparison with Image and Video Editing Baselines on Out-of-Domain Testing Data (Part 2 of 2).}
  Our method harmonizes color tone and synthesizes realistic lighting and shadows, while editing baselines often fail to produce physically plausible shadowing. Although both can reduce reconstruction artifacts, baselines tend to hallucinate inconsistent content and over-edit well-reconstructed regions, whereas our method preserves scene geometry and input structure. Moreover, image-editing baselines introduce frame-to-frame jitter, whereas our model maintains strong temporal coherence.
  }
  \label{fig:qualitative_supp_part2}
  \vspace{-4mm}
\end{figure*}

We provide additional qualitative comparisons with baselines in Figures~\ref{fig:qualitative_supp_part1} and \ref{fig:qualitative_supp_part2}, on out-of-domain Waymo scenes. Our model reliably harmonizes color tone and synthesizes scene-consistent lighting and shadows, whereas state-of-the-art image and video editing baselines often fail to generate physically plausible shadowing for inserted objects. Although these baselines can partially reduce reconstruction artifacts, they frequently hallucinate content or modify regions that should remain intact. In contrast, our approach preserves scene geometry and underlying input structures while maintaining strong temporal consistency across adjacent frames.

\begin{figure*}[t!]
  \centering
  \vspace{-3mm}
   \includegraphics[width=0.8\linewidth]{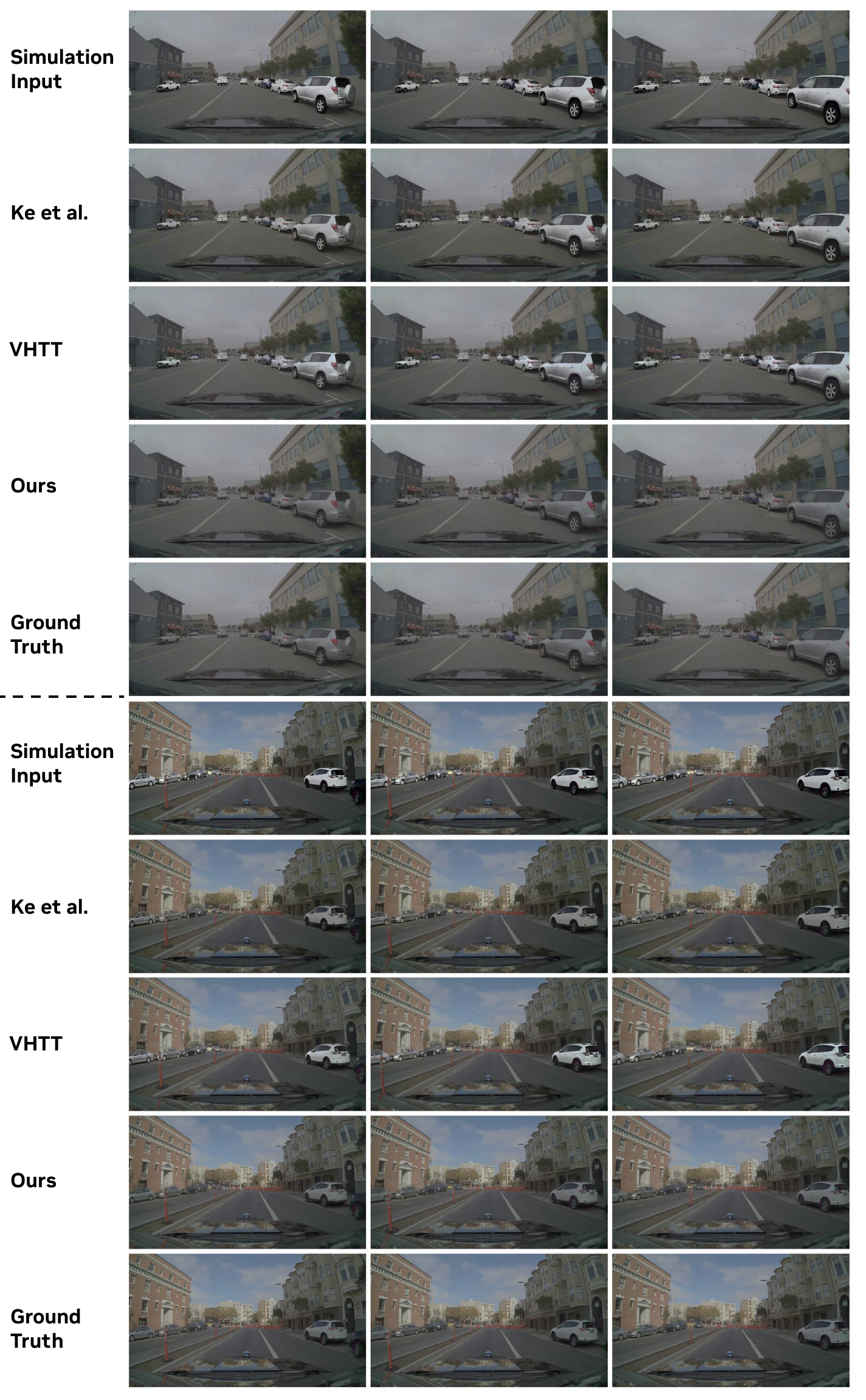}\\
\vspace{-4mm}
\caption{\footnotesize
\textbf{Additional Comparison with Video Harmonization Baselines on ISP modification held out test set.}}
\vspace{-4mm}
   \label{fig:qualitative_supp_videoharm}
   \vspace{0mm}
\end{figure*}

In Figures~\ref{fig:qualitative_supp_videoharm} and \ref{fig:qualitative_supp_videoharm2}, we further provide additional comparisons with specialized harmonization methods. While these approaches adjust foreground color appearance, they do not synthesize realistic shadows and cannot address reconstruction artifacts by design, limiting their applicability in simulation-enhancement pipelines.

\subsection{Additional Qualitative Comparison with Ground-Truth Labels} 
\begin{figure*}[t]
    \centering
    \includegraphics[width=\linewidth]{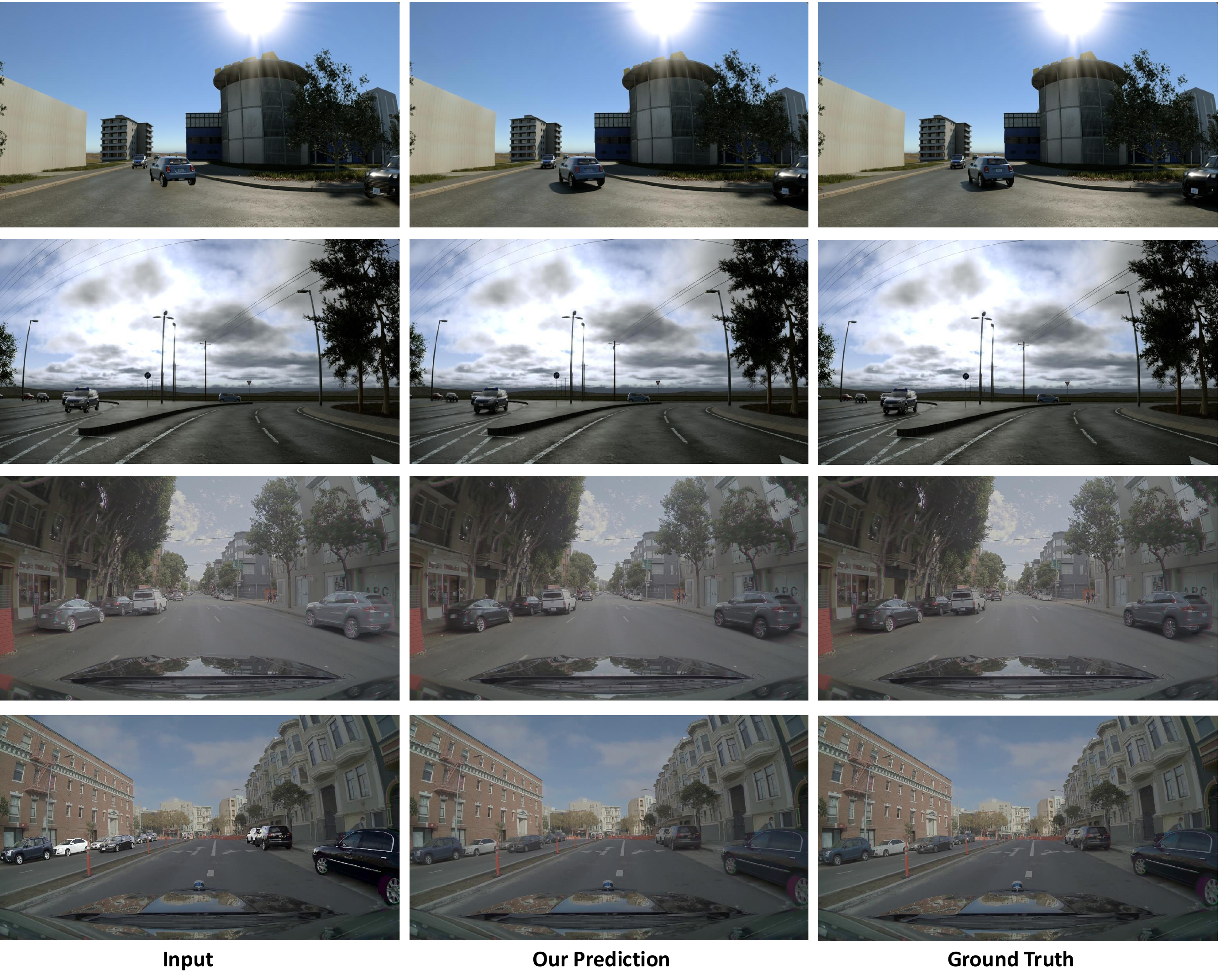}
    \vspace{-6mm}
    \caption{\footnotesize \textbf{Comparison with Ground Truth on Holdout Datasets.} Our model’s predictions closely match the ground-truth real-world captures, producing faithful, physically plausible results suitable for online simulation systems.}
    \label{fig:gt_comparison}
    \vspace{-0mm}
\end{figure*}

We additionally present qualitative results on our holdout datasets and compare the predicted outputs against the corresponding ground-truth labels. As shown in \cref{fig:gt_comparison}, our model's predictions closely match real-world scenarios, illustrating that it produces physically faithful enhancements suitable for online deployment in simulation pipelines.

\vspace{-0mm}

\vspace{-0mm}

\clearpage

\vspace{-0mm}

\vspace{-0mm}

\end{document}